\documentclass{article}

\PassOptionsToPackage{numbers, compress}{natbib} 
    \usepackage[preprint]{neurips_2024}

\usepackage[utf8]{inputenc} 
\usepackage[T1]{fontenc}    
\usepackage{hyperref}       
\usepackage{url}            
\usepackage{booktabs}       
\usepackage{amsfonts}       
\usepackage{nicefrac}       
\usepackage{microtype}      
\usepackage[dvipsnames]{xcolor}
\usepackage{graphicx}
\usepackage{lipsum}  
\usepackage{subcaption}
\usepackage{adjustbox}
\usepackage{wrapfig}
\usepackage{multirow} 
\usepackage{amsmath}
\usepackage{enumitem}
\usepackage{amssymb}
\usepackage{float}

\DeclareMathOperator*{\argmax}{arg\,max}
\DeclareMathOperator*{\argmin}{arg\,min}

\usepackage{enumitem}
\setlist{topsep=0pt, leftmargin=*}
\hypersetup{hidelinks}

\renewcommand{\paragraph}[1]{{\bf #1}~~}

\newcommand{\meanstd}[2]{#1 \!{\scriptsize $\pm$\! #2}}
\newcommand{\resultssep}{/}

\newcommand{\graph}{\mathcal{G}} 
\newcommand{\task}{\mathcal{T}}
\newcommand{\taski}{\mathcal{T}_i}
\newcommand{\taskti}{\task^{(t)}_i}
\newcommand{\taskt}{\task^{(t)}}

\newcommand{\agents}{A}
\newcommand{\agent}{\mathrm{a}}
\newcommand{\agenti}{\agent_i}
\newcommand{\expectation}{\mathbb{E}}

\newcommand{\etal}{et al.}

\newcommand{\red}[1]{\textcolor{red}{\textrm{#1}}}
\newcommand{\blue}[1]{\textcolor{blue}{\textrm{#1}}}

\usepackage[normalem]{ulem}

\title{%
Distributed Continual Learning
}

\author{%
  Long Le, Marcel Hussing, and Eric Eaton \\
  Department of Computer and Information Science \\
  University of Pennsylvania \\
  Philadelphia, PA 19104 \\
  \texttt{\{vlongle, mhussing, eeaton\}@upenn.edu} \\
}

\begin{document}

\maketitle
\begin{abstract}

This work studies the intersection of continual and federated learning, in which independent agents face unique tasks in their environments and incrementally develop and share knowledge. We introduce a mathematical framework capturing the essential aspects of distributed continual learning, including agent model and statistical heterogeneity, continual distribution shift, network topology, and communication constraints. Operating on the thesis that distributed continual learning enhances individual agent performance over single-agent learning, we identify three modes of information exchange: data instances, full model parameters, and modular (partial) model parameters. We develop algorithms for each sharing mode and conduct extensive empirical investigations across various datasets, topology structures, and communication limits. Our findings reveal three key insights: sharing parameters is more efficient than sharing data as tasks become more complex; modular parameter sharing yields the best performance while minimizing communication costs; and combining sharing modes can cumulatively improve performance. 
\end{abstract}

\section{Introduction}

Throughout human history, collective intelligence has driven significant scientific and technological advancements \cite{pmid21357237, Fagan2012-FAGCSK}. The pooling and exchange of knowledge among individuals have often led to outcomes that far exceed the capabilities of isolated efforts. This collective principle is now reflected in contemporary machine learning research, where combining data \cite{levine2016learning, embodimentcollaboration2024open} and computational resources \cite{chowdhery2022palm,fedus2022switch} can produce significantly more capable models. However, training such large-scale foundation models typically requires a high degrees of coordination among agents (e.g., through a central server), and is only done \textbf{once} due to costs.

In contrast, we envision a future of ubiquitous embodied agents in the wild, such as vehicles in different cities or service robots in various homes.~While foundation models provide a strong starting point, these robots need to \textbf{continually} learn new skills for their specific contexts. We posit that the selective sharing of knowledge among these agents through a communication network can substantially complement and enhance individual learning \cite{Soltoggio2024}. For example, consider a fleet of household robots deployed in homes across the nation. Commonalities across house layout, possessions, user preferences, etc.~would enable one robot to benefit from the experience of others, improving collective performance and accelerating learning, yet individual aspects of each home and owner still require personalization of each robot.
Despite its potential, this domain presents several key challenges inherited from related fields, including:
\vspace{-0.5em}
\begin{itemize}[nosep]
    \item \textbf{Agent Heterogeneity:} The diversity in agent functional forms, such as varying neural network architectures and learning algorithms, as well as differing local data and tasks, necessitate consideration of what form of knowledge to share and under what assumptions about agent commonality.
    \item \textbf{Continual Adaptation:} Algorithms for learning and sharing must facilitate ongoing skill acquisition and adaptation while retaining previously learned knowledge.
    \item \textbf{Selective Transfer:} Constraints on communication budgets and frequency necessitate efficient and selective knowledge transfer.
    \item \textbf{Arbitrary Topologies:} Sharing algorithms must be distributed and able to operate effectively in arbitrary communication topologies.
\end{itemize}

To address these challenges, we first formalize a mathematical framework for this problem of distributed continual learning (DCL), which encapsulates the essential aspects of this emerging field \cite{Soltoggio2024}. Within this framework, we identify three fundamental modes of knowledge exchange: data instances, full model parameters, and partial (modular) parameter sharing, each with its own trade-offs and benefits. We then develop selective methods tailored to each mode and conduct extensive experiments to highlight their strengths across datasets, communications topologies, and constraints.

Our contributions are as follows:
\vspace{-0.5em}
\begin{itemize}[nosep]
    \item \textbf{Comprehensive Framework:} We present a mathematical framework for distributed continual learning (DCL) that integrates features from several related fields.
    \item \textbf{Selective Algorithms:} We develop and analyze algorithms for three distinct modes of sharing: data instances, full model parameters, and partial (modular) parameter sharing.
    \item \textbf{Empirical Insights:} Our extensive empirical studies reveal that (1) parameter sharing is more efficient as tasks become more complex, (2) modular parameter sharing provides the best performance with minimal communication, and (3) combining sharing modes inherits the benefits of each mode.    
    \item \textbf{Baselines and Evaluation Protocols:} We provide competitive sharing baselines and emphasize the importance of a robust evaluation protocol for future research in DCL, taking into account the hidden cost of communication and realistic considerations, such as agent model heterogeneity.
\end{itemize}

\section{Related Work} \label{sec:related_work}
The problem of distributed continual learning is related to several other topics, overviewed below:

\paragraph{Federated learning}
Traditional federated learning \cite{mcmahan2023communicationefficient, Li_2020} focuses on training one global model using i.i.d data distributed across edge devices. More recent methods address challenges associated with non-i.i.d data among clients \cite{li2020federated, Liang_2022, li2021fedbn, wang2022personalized, ma2021heterogeneous, stich2019local}, continual learning \cite{yoon2021federated, shenaj2023asynchronous}, 
and decentralized architectures \cite{sun2021decentralized, vogels2020powergossip}. These works typically compare themselves against other federated methods. Our work extends these concepts by incorporating both continual and distributed learning within a federated framework and comparing federated learning against other modes of sharing.

\paragraph{Continual Learning}
Continual learning (CL) \cite{pmlr-v28-ruvolo13, 10.5555/3276462, PARISI201954} aims to develop models capable of learning new tasks without forgetting the knowledge of past tasks. To combat catastrophic forgetting \cite{MCCLOSKEY1989109}, methods have used regularization to limit drastic changes to network parameters during successive learning \cite{Kirkpatrick_2017,zenke2017continual,li2017learning}, rehearsal strategies that maintain a small subset of past data for periodic retraining \cite{doi:10.1080/09540099550039318,isele2018selective, lopezpaz2022gradient}, preservation of critical network nodes and pathways \cite{Jung2020Adaptive}, and architectural modification strategies to add parameters and capacity for new tasks \cite{schwarz2018progress, yoon2020scalable, rusu2022progressive, yoon2018lifelong}. Recent work has shown the benefits of using modular  representations for architectural modification \cite{meyerson2018shared, kirsch2018modular, mendez2023reuse,ijcai2019p393, mendez2022composuite}. These methods decompose a neural network into independent and reusable knowledge components, for example, through neural block modules \cite{meyerson2018shared} or tensor factorization \cite{ijcai2019p393}. In a CL setting, modular representations allow for the selective reuse of prior knowledge and expansion as required  \cite{yoon2020scalable, mendez2021lifelong, mendez2022modular, mendez2023reuse}. Modular learning is particularly relevant to this work as it provides a principled way to organize learned information into reusable components that can be shared between agents.

\paragraph{Multi-agent Systems}
Research on distributed and multi-agent systems \cite{8352646, Stone2000, 10.1007/978-3-031-21671-8_10, le2024articulate} spans several decades, focusing on applications in planning \cite{deWeerdt2009, Durfee2001}, control \cite{piccoli2023control, 8755265, le2021multi, foerster2017counterfactual}, and game theory \cite{5160273}. Unlike our work, these systems often aim to solve a single shared task or manage agent behavior \cite{Stone2000}. Nonetheless, considerations in multi-agent systems including network topology, communication constrain, and coordination requirements help to inform our formalism. 


\paragraph{Distributed Continual Learning} Relatively few works have explored continual learning in distributed or multi-agent settings.  
An early work in distributed continual learning by Rostami et al.~\cite{rostami2018multiagent} tackles the problem using shallow factorized models, framing it as distributed dictionary optimization which requires that neighboring agents have similar knowledge bases. Subsequent work \cite{8918888} extends this approach by allowing each agent to maintain some local expertise in addition to globally shared knowledge. However, this type of sharing is not selective as it fails to consider the transferability of knowledge between agents, and more importantly, it is restricted to simple parametric models (e.g., logistic regression) and thus not suitable for modern neural networks. More recent approaches allow task-specific transfer of knowledge via modulating masks \cite{nath2023sharing} or lightweight heads \cite{ge2023lightweight}. However, these methods require all agents to share the same network backbone. Our modular sharing, on the other hand, affords the flexibility of model heterogeneity between agents.
Furthermore, these works typically compare themselves against single-agent baselines instead of other sharing modes, and completely ignore the hidden cost of communication and other practical considerations including model heterogeneity. Our work aims to provide more competitive baselines, and more rigorous evaluation for future DCL research.

\section{Distributed Continual Learning Framework} \label{sec:framework}
We now formalize the idea of distributed continual learning (DCL), with an emphasis on practical considerations such as supporting heterogeneous agents and communication constraints.  Related works on DCL, described above, have explored variations of the problem; here we specify a framework that integrates and formalizes ideas from these works to unify them. 
Our framework consists of a directed graph $\graph = \{A,E\}$ of $|A|$ nodes (agents) and $|E|$ edges that represent communication connections between agents.  The collective of agents $\agents$ face a set of tasks $\mathcal{T}$, where each task $\taskt$ is a supervised learning problem with data $X^{(t)} \subseteq \mathcal{X}^{(t)}$ and labels $Y^{(t)} \subseteq \mathcal{Y}^{(t)}$ drawn from some data distribution for that task $\{\mathcal{X}^{(t)}, \mathcal{Y}^{(t)}\}\sim\taskt$. Each agent $\agenti$ has a local view over this task space and faces a subset $\taski \subseteq \mathcal{T}$ of these tasks. For example, robots in home versus commercial kitchens would share some tasks, but other tasks differ between the settings. Note that $\{\taski\}_{i=1}^{|A|}$ covers $\task$. 

Rather than learning all tasks en masse, each agent faces the tasks $\taski$ in series: $\taski^{(1)}, \taski^{(2)},\ldots,\taskti,\ldots$, over its lifetime. Agent $\agenti$ learning task $\taskti$ corresponds to estimating parameters $\theta_i^{(t)}$ for a model $f_{\theta_i}^{(t)}: \mathcal{X}^{(t)} \mapsto \mathcal{Y}^{(t)}$ by solving a risk minimization problem with loss $\mathcal{L}_i^{(t)}\big(f^{(t)}_{\theta_i}(X^{(t)}), Y^{(t)}\big)$.  Once task $\taskt \in \taski$ is seen by 
the collective, agent $\agenti$ may be evaluated on it at any time, and so agents must retain knowledge of previous tasks. For two agents $\agenti$ and $\agent_j$, note that their respective task sets $\taski$ and $\task_j$ are not necessarily disjoint, and so may or may not overlap and share identical or related tasks. The task similarity $S(\task_i, \task_j)$ can be characterized using any measures of statistical difference between distributions, e.g., KL divergence \cite{10.1214/aop/1176996454}. 

The goal is for the agent collective to learn models for all tasks; since tasks may overlap or be related between agents, there is incentive for them to maximize performance by transferring knowledge from their previous tasks and across agents. Through the network $\graph$, agent $\mathrm{a_i}$ can find other agents such as $\mathrm{a_j}$ who share similar tasks, allowing them to learn from each other. Each edge $e_{ij}$ represents a communication connection between agents $\agenti$ and $\agent_j$, and, for realism, has a set budget $\mathtt{b}_{ij}$ of the maximum number of tokens allowed per communication and a permitted communication frequency $\mathtt{f}_{ij}$.  The overall objective of the multi-agent collective can therefore be given by the cumulative expected loss of all agents over their task distributions, given the knowledge they are able to share: 
\begin{equation}
    \begin{split}
    \min_{\Theta_1, \ldots \Theta_{|\!A\!|}}
    \sum_{\taskt \in \task} \sum_{\agenti \in \agents} \expectation_{(X^{(t)},Y^{(t)}) \sim \taskt} \ \Pr(\taskt \!\mid \task_i) \cdot \mathcal{L}_i^{(t)}\!\left(f^{(t)}_{\theta_i}(X^{(t)}), Y^{(t)} \mid \mathcal{K}_i \right)\\
    \mbox{s.t. } \mathcal{K}_i = \textstyle{\bigcup_{j=1}^{|A|} K_{j,i}} 
    \mbox{ and }\mathrm{size}(K_{j,i}) \leq \mathrm{b}_{ji} \cdot \mathrm{f}_{ji} \cdot C \ \ \forall \{j | 1 \leq j \leq |A|, j \neq i\} \enspace , 
    \end{split} \label{eq:CollectiveObjective}
\end{equation}
where $\Theta_i$ is all learned parameters for agent $\agenti$, $\mathcal{K}_i$ represents the set of knowledge $\agent_i$ accumulated from other agents, $K_{j,i}$ is the knowledge sent from $\agent_j$ to $\agent_i$ (with $K_{i,i} = \emptyset$), and $C$ is a global clock counter for all communications that starts at 0 and increments each time agents receive a task. 

Consequently, because the amount of transfer between agents is constrained by the communication bandwidth and frequency in Equation~\ref{eq:CollectiveObjective}, the agents necessarily must be efficient in transferring knowledge to maximize collective performance. Since tasks are experienced sequentially by the agents, the ability to optimize this objective improves over time, as $C$ increases. Note also that this objective includes zero-shot performance on tasks that are unseen by an agent, but those unseen tasks may or may not be encountered by other agents in the collective.

This framework features several key characteristics that make this a challenging and realistic problem:\vspace{-0.5em}
\begin{itemize}[nosep]
\item \textbf{Model heterogeneity}: Agents might be functionally different (e.g., neural network architectures, learning methods, etc.) Compatibility is assumed only through the information $K_{ij}$ transferred.
\item \textbf{Task and data heterogeneity}: Agents observe task distributions $\taski$ and data that are local.
\item \textbf{Continual learning}: Agents must solve new tasks over time, necessitating the adaptation of the $\Theta_i$'s.  Agents may be evaluated on any task at any time, necessitating retention. The framework could support task- or class-incremental learning \cite{vanDeVen2019Three}, depending on how the tasks $\task$ are formulated.
\item \textbf{Collaborative transfer}: Collaboration provides a mechanism for agents to share information to improve collective performance, and could be either sender- or receiver-initiated. 
\item \textbf{Communication constraints}: The framework enforces realistic constraints on communication bandwidth $\mathtt{b}$ and frequency $\mathtt{f}$, which may differ by application. For example, embedded devices may send frequent but small updates (high $\mathtt{f}$, low $\mathtt{b}$), while satellites may limit both due to power.
\item \textbf{Arbitrary topology}: Unlike the centralized server-client model of traditional federated learning, our framework supports agents connected in an arbitrary graph, potentially based on geolocation.
\end{itemize}

This framework unifies a variety of existing works on related problems, which can be specified as various assumptions on optimizing Eq.~\ref{eq:CollectiveObjective}.  For example, single-agent continual learning optimizes the $\mathcal{L}_i^{(t)}$'s independently under the assumption that $\mathcal{K}_i = \emptyset$.  Rostami~\etal~\cite{rostami2018multiagent} and Mohammadi and Kolouri~\cite{8918888} explore a variant assuming a factorized model for $K$ and unlimited communication. Ge~\etal~\cite{ge2023lightweight} assume a static shared backbone among all agents and share task-specific heads, biases, and task clusters as $K$, again under unlimited communication.

\section{Modes of Knowledge Sharing} \label{sec:method}
Prior works in continual \cite{pmlr-v28-ruvolo13, 10.5555/3276462, PARISI201954} and distributed learning \cite{rostami2018multiagent,8918888,nath2023sharing, ge2023lightweight} have explored some aspects of how to optimize Eq.~\ref{eq:CollectiveObjective} but largely ignored the form and constraint on $\mathcal{K}_i$. To bridge this gap, we focus on this fundamental challenge within Eq.~\ref{eq:CollectiveObjective} that has been under-investigated: {\em what information should agents share in order to maximize collective performance?} The communication constraints and heterogeneity among agents and tasks creates obstacles for addressing this challenge, including limiting the forms (modes) of information that can be conveyed. We study this problem by considering sharing of information at three different levels: instance-based, full-model, and partial-model. We first describe these modes, and then propose solutions to DCL with each.

{\bf Instance-based sharing of {\em data}} (i.e., training examples) is the most generic mode, since it is model-agnostic and only requires agents to be compatible in their input data formats. The downside is that each data instance contains very limited information. So, this mode has low communication efficiency in general, although it may be efficient when only a few specific examples need to be shared. 

{\bf Full-model sharing of {\em full model parameters}} provides an efficient mechanism to share complete information among agents. Compression schemes could make transmission even more efficient, but the information content of the exchange would remain approximately the same. However, it also imparts strong assumptions that the agents' models are directly compatible and that their task distributions are likely similar at any given time. This is most similar to federated methods that attempt to learn a unified model for all tasks in a distributed fashion.

{\bf Partial-model sharing of reusable {\em model components}}  represents a compromise between low-level data sharing and full model sharing. It relaxes assumptions on the agents' model compatibility, requiring that they only be capable of utilizing the components within their model architecture, with that architecture and learning methods able to differ among agents. It also reduces the assumptions about common task distributions, allowing for selective sharing based on the similarity of the current task to some tasks in the past encountered by the collective. Although different forms of partial-model transfer are available that have been used for CL, e.g., factorized representations \cite{rostami2018multiagent,8918888}, we impose an additional requirement---that the module be self-contained, so that they can be transferred independently. In contrast, factorized representations often have interdependencies  (e.g., among model basis vectors), complicating the ability to share just one factor between agents. For this reason, we focus on transfer of self-contained modules, and in particular, modules that are compositional \cite{mendez2021lifelong} in nature, thereby avoiding side effects from different combinations of modules.

\subsection{Data Sharing}

To determine what data to share in DCL, we explore a receiver-initiated approach, similar to active learning~\cite{settles2009active}, where the receiving agent identifies gaps in its knowledge base and puts out requests for appropriate data from neighboring agents. Previous work has also explored sender-initiated models, such as training a sharing model that predicts the utility of an agent's data to other agents based on estimated novelty~\cite{pmlr-v164-geng22a}. Such sender-initiated methods assume the usefulness of data to other agents, rather than letting demand drive sharing, and hinge on frequent communication to train the sharing model. Conversely, other agents may have broader information on the data than an individual agent, and so sender-initiated models may fill gaps of which receiver agents were unaware.

In our receiver-initiated data sharing, $\mathtt{Recv}$, when agent $\agenti$ is learning task $\taskt$, it identifies $q$ difficult instances and requests more data that is similar. To identify the $\mathtt{q}$ hardest instances, we compute the cross entropy loss on the validation set $\mathcal{V}_i^{(1:t)}$ of all tasks encountered so far: $\mathcal{Q}_i^{(t)} = \argmax_{I \subseteq \mathcal{V}_i^{(1:t)}} \sum_{(x, y) \in I} \mathcal{L}_{CE}(f_{\theta_i}(x), y) \mbox{ s.t. } |I| = \mathtt{q}$.
Agent $\mathrm{a}_i$ then queries its neighbors $ \mathcal{N}_i$ to return the $\mathtt{k}$  nearest instances to each query in  $\mathcal{Q}_i^{(t)}$, based on a distance measure $d(\cdot)$. For $d(\cdot)$, we use the cosine distance of the latent representation from the penultimate layer of the neural network model $ \phi_{\theta_i}(x)$. The result $\mathcal{R}_{j}^{(t)}(x)$ to a query $x$ computed by a sender $\agent_j$ is the $k$ nearest instances in the sender database $\mathcal{\tilde{D}}_j^{(1:t)}$: $\mathcal{R}_{j}^{(t)}(x) = \argmin_{R \subseteq \mathcal{\tilde{D}}_j^{(1:t)}} \sum_{(x', y') \in R} d(\phi_{\theta_j}(x), \phi_{\theta_j}(x')) \mbox{ s.t. } |R| = \mathtt{k}$, where database $\mathcal{\tilde{D}}_j^{(1:t)}$ represents a select subset of $\agent_j$'s training data from known tasks, such as stored in a replay buffer to avoid catastrophic forgetting \cite{isele2018selective}.  
 The values of $\mathtt{q}$ and $\mathtt{k}$ are determined by the communication limit $\mathtt{b}$ of the network such that $\mathtt{b} = \mathtt{q} \mathtt{k}.$

We also implement a simpler sharing mechanism, $\mathtt{Simp}$, where the receiver requests data from certain classes, and the sender samples data instances uniformly from that class from its local database.
In $\mathtt{Simp}$, the receiver selects classes $ \mathcal{C}_i \subseteq \mathcal{Y}^{(1:t)} $ 
and computes the worth of each class using the cross entropy loss $\mathcal{L}_{CE}$ on the validation set $\mathcal{V}^{(1:t)}$. Let $ \mathcal{W}_i $ represent the worth of each class, calculated as $ \mathcal{W}_i(c) = \mathcal{L}_{CE}(c) $ for class $ c $. The receiver then requests $ \mathtt{b} $ instances from the sender.
Given a set of requested classes $ \mathcal{C}_i $, the sender $\agent_j$ computes the intersection of requested classes with its available classes and normalizes the instances it will send based on the worth values.
Let $ \mathcal{C}_j \subseteq \mathcal{Y}^{(1:t)} $ be the set of classes available at the sender. The sender computes the intersection $ \mathcal{C}_{i,j} = \mathcal{C}_i \cap \mathcal{C}_j $ and samples data instances uniformly from these classes: $\mathcal{S}_i^{(t)} = \bigcup_{c \in \mathcal{C}_{i,j}} \bigl\{ (x, y) \in \mathcal{D}_j^{(1:t)} \mid y = c \bigr\}$. 
The sender normalizes the number of instances of each class $ c $ by the worth value $ \mathcal{W}_i(c) $, ensuring the total number of instances sent does not exceed the communication budget $ \mathtt{b} $: $\mathbf{N}_c = \Bigl\lfloor \frac{\mathcal{W}_i(c)}{\sum_{c' \in \mathcal{C}_{i,j}} \mathcal{W}_i(c')} \cdot \mathtt{b} \Bigr\rfloor$.
Here, $ \mathbf{N}_c $ is the number of instances of class $ c $ to be sent, and $ \mathtt{b} $ is the total communication budget. The sender then sends $ \mathbf{N}_c $ instances for each class $ c \in \mathcal{C}_{i,j} $.

\subsection{Full Model Parameter Sharing} \label{sub:federated}
To share full model parameters, we leverage federated learning techniques, considering three methods: federated averaging (FedAvg) \cite{rothchild2020fetchsgd}, FedProx \cite{li2020federated}, and FedCurv \cite{shoham2019overcoming}. These methods were chosen since they are popular baselines used in most related work surveyed in Sect.~\ref{sec:related_work}, and do not rely on additional assumptions (e.g., the ability to pretrain on fractal images as in Shenaj~\etal~\cite{shenaj2023asynchronous}), making them directly comparable to our isolated learning baselines. 

FedAvg is a well-established method where local models are averaged to create a global model. FedProx addresses statistical heterogeneity by encouraging local models to stay close to the global model via L2 regularization. Instead of uniformly penalizing deviations in the weights of the neural network, FedCurv utilizes the diagonals of Fisher information matrices to prioritize preserving parameters important to previous tasks, similar to EWC \cite{Kirkpatrick_2017}. However, FedCurv requires each node to send over its Fisher diagonals, doubling the bandwidth, and constrains the optimization of one node to preserve the knowledge of other nodes. Furthermore, both FedProx and FedCurv modify the local optimization step of FedAvg. To investigate the effect of improving the aggregation step instead, we introduce FedFish to utilize the Fisher diagonals during aggregation at round $\tau$ when learning task $t$: $\theta_i^{(\tau+1)} = \tilde{\text{diag}}(I_i) \cdot \theta_i^{(\tau)} + (1 - \tilde{\text{diag}}(I_i)) \cdot \frac{1}{N} \sum_{j=1}^N \theta_j^{(\tau)}$, 
where $\tilde{\text{diag}}$ is the Fisher diagonals capturing how important each parameter is to previous tasks $\task_{1:t}$ normalized by a softmax probability distribution to be between $[0, 1]$. This aggregation rule prioritizes averaging less important local parameters ($\tilde{\text{diag}} \sim 0$) while keeping the important ones ($\tilde{\text{diag}} \sim 1$) close to the agent's previous copy. Agent $\agenti$ only uses its own Fisher diagonals so the bandwidth requirement is half of FedCurv. We implement all four of these federated methods to allow exchanging information between two connected nodes in a distributed network without relying on a central coordinator.



\subsection{Modular Parameter Sharing} \label{sub:modular}
For partial-model sharing, we build upon the compositional continual learning method by
Mendez and Eaton \cite{mendez2021lifelong}, which learns a knowledge base of reusable modules (e.g., mini neural networks) that are then composed together into task models. 
During training, new modules are considered through a process called component dropout, and added to the knowledge base only if needed to solve the task adequately.
In their single-agent setting, new candidate modules are initialized randomly; our multi-agent collective opens the possibility of stronger initialization using relevant modules from other agents, potentially boosting initial learning.

To adapt this mechanism to DCL, given a fixed communication budget $\mathtt{b}$, each agent $\agenti$ may share $k$ modules with its neighbors. Agent $\agenti$ ranks its available modules on their benefit to other agents, based on transferrability estimation between the sender $\agenti$'s (known) tasks and receiver agent $\agent_j$'s (new) tasks. We employ two transferrability methods: the Log Expected Empirical Prediction (LEEP) metric \cite{nguyen2020leep}  and a cruder intersection-over-union (IoU).
Given $\agenti$'s model $f_{\theta_i}^{(t)}$ trained on a source task with label set $\mathcal{Y}_i^{(t)}$, and a target task with label set $\mathcal{Y}_j^{(t)}$, the LEEP score is: $\text{LEEP}(f_{\theta_i}^{(t)}, \mathcal{V}^{(t)}_j) = \frac{1}{|\mathcal{V}^{(t)}_j|} \sum_{(x_j, y_j) \in \mathcal{V}^{(t)}_j} \log \left( \sum_{y_i \in \mathcal{Y}_i^{(t)}} P(y_i | x_j; f_{\theta_i}^{(t)}) P(y_j | y_i) \right)$, 
where $ P(y_i | x_j; f_{\theta_i}^{(t)}) $ is the softmax output of the source model for the target input $x_j$, and $P(y_j | y_i)$ is estimated using the empirical distribution of the target labels given the source labels.
The intersection-over-union (IoU) estimate of task similarity is computed as: $\text{IoU}(\mathcal{Y}_i, \mathcal{Y}_j) = \frac{|\mathcal{Y}_i \cap \mathcal{Y}_j|}{|\mathcal{Y}_i \cup \mathcal{Y}_j|}$. 
Although we focus on LEEP and IoU, other transferability methods could be used instead, such as OTCE \cite{Tan_2021_CVPR}, F-relatedness \cite{10.1007/978-3-540-45167-9_41}, A-distance \cite{kifer2004}, or discrepancy distance \cite{mansour2023domain}. Agent $\agenti$ selects the $k$ modules from tasks with the highest transferrability estimates that also have new modules added due to component dropout, ensuring these modules encode information about the relevant tasks and then shares with $\agent_j$. We refer to this modular sharing method as $\mathtt{modmod}$.

From the receiver agent $\agent_j$'s perspective, it must choose a module among those shared from its neighbors to initialize its new module in the upcoming task. One approach is to optimize each module for a short period and select the best-performing one ($\mathtt{TryOut}$ strategy). To reduce the GPU computation and memory cost of a na\"{i}ve implementation of $\mathtt{TryOut}$, we extend the $\mathtt{ComponentDropout}$ procedure of Mendez and Eaton~\cite{mendez2021lifelong} to optimize multiple candidate modules in a round-robin fashion.  Alternatively, the receiver can trust the task similarity estimates from the neighbors and select the module that has the highest score reported by the sender ($\mathtt{TrustMetric}$ strategy).

\section{Experiments}
We compare the strategies from Sect.~\ref{sec:method} on different datasets, communication constraints, and network topologies. We use four common datasets (MNIST, KMNIST, FashionMNIST, and CIFAR-100) in a continual learning setting, creating tasks following prior work \cite{mendez2021lifelong}. To explore task heterogeneity, we also create a $\mathtt{combined}$ setting that includes tasks from MNIST, KMNIST, and FashionMNIST. In this setting, all agents receive a mixture of a few initial tasks from the three source datasets; then, they segregate into three roughly equal-sized groups, each receiving tasks drawing from exactly one source dataset. Note that information can still potentially flow from, e.g. MNIST agents to FashionMNIST agents, provided that they are connected in the network. Tasks are learned consecutively in a continual learning setting; during the course of training on task $\task^{(t)}$, we compute the average accuracy over all tasks seen so far $\task^{(1)},...,\task^{(t)}$ on their test sets every 10 epochs. Learning curves are averaged across all tasks, agents, and random seeds. Complete implementations and experiment scripts are available at: {\tt https://github.com/OMITTED\_FOR\_BLIND\_REVIEW}.

We consider two forms of neural network models: {\em monolithic} (i.e., standard feed-forward) neural networks or {\em modular} neural networks \cite{mendez2021lifelong}. Experience replay is employed to mitigate catastrophic forgetting.  As a lower bound, we compare against single-agent isolated learning as a $\mathtt{baseline}$. For data sharing, we use $\mathtt{Recv}$ for MNIST variants and the $\mathtt{combined}$ dataset, and use $\mathtt{Simp}$ for CIFAR-100. For full model parameter sharing, we implement $\mathtt{FedAvg}$~\citep{rothchild2020fetchsgd}, $\mathtt{FedProx}$~\citep{li2020federated}, $\mathtt{FedCurv}$~\citep{shoham2019overcoming}, and $\mathtt{FedFish}$. For $\mathtt{modmod}$, we use use the $\mathtt{LEEP}$ metric for the $\mathtt{combined}$ dataset and $\mathtt{IoU}$ for the others. $\mathtt{TrustSim}$ module selection is used in the basic experiment (Sec. \ref{sub:monolithic}-\ref{sub:modularExps}) but as we get more candidate modules in the increasing budget evaluation (Sec. \ref{sub:comm_constraints}) we use $\mathtt{TryOut}$ instead.
Additional details and justifications for these choices can be found in the Appendix.

\subsection{What to share in a monolithic distributed setting?} \label{sub:monolithic}

\begin{figure}[h]
  \centering
  \includegraphics[width=\textwidth]{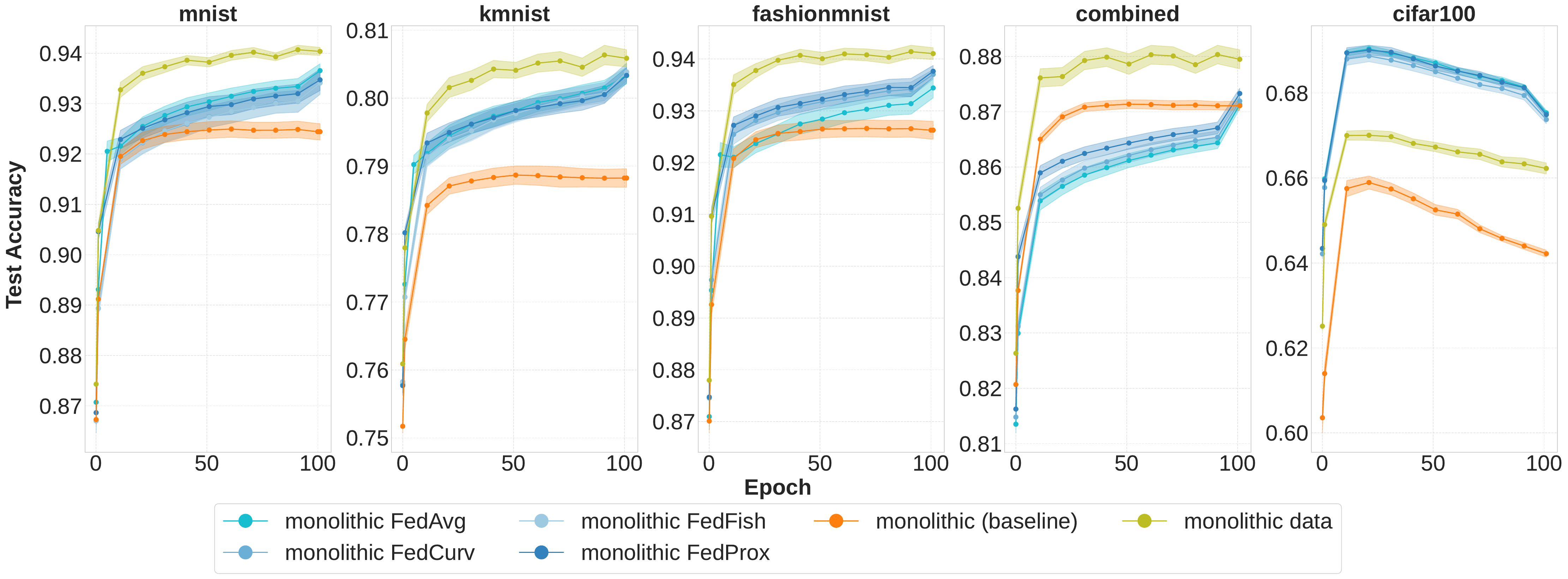}
    \caption{Mean test accuracy and standard error of the no-sharing baseline, data sharing, and federated learning with monolithic models. Sharing data is best for easier tasks while federated learning is best for harder tasks (i.e., CIFAR-100). In the heterogeneous $\mathtt{combined}$ dataset, where agents face very different tasks, aggregating models via federated methods is worse than single-agent learning.}
  \label{fig:monolithic_lc}
\end{figure}

Our first experiments examine performance of DCL using monolithic (standard) neural networks. Fig.~\ref{fig:monolithic_lc} shows that sharing data in this DCL setting outperforms federated learning methods and the baseline across all datasets, except CIFAR-100 where federated methods are the best. This is because a harder dataset like CIFAR-100 requires more training data to achieve good performance, diminishing the relative value of each data instance compared to model parameters. In a more extensive experiment detailed in Sect.~\ref{sub:comm_constraints}, we find that increasing the data budget eventually allows data sharing to surpass federated methods in CIFAR-100.

There is little difference observed between different federated learning methods. The $\mathtt{combined}$ dataset, characterized by a high degree of heterogeneity between agents, is the only setting where federated learning is outperformed by the baseline.

\subsection{What to share in a modular distributed setting?} \label{sub:modularExps} 

Prior work \cite{Kirkpatrick_2017} has noted the challenge of catastrophic forgetting that monolithic networks face in dealing with long-horizon continual problems. Modular networks \cite{mendez2023reuse,mendez2021lifelong} have been proposed as a solution and have shown superior performance in maintaining task accuracy over time. We now examine the effect of different sharing modes on the DCL problem using modular networks.  With the switch to modular networks, we also can now compare against partial sharing using $\mathtt{modmod}$.

\begin{figure}[h]
  \centering
  \includegraphics[width=\textwidth]{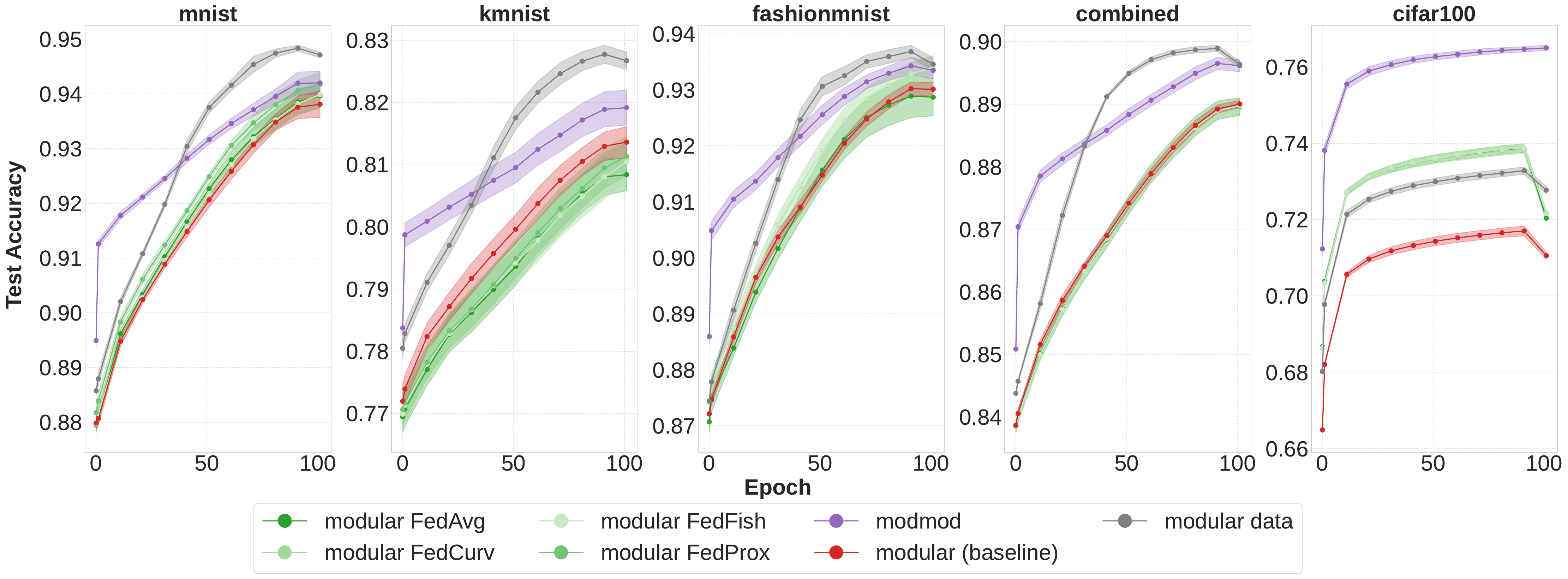}
    \caption{Mean test accuracy and standard error of the no-sharing baseline, data sharing, full-model sharing (federated learning), and  partial model sharing ($\mathtt{modmod}$) with modular models. $\mathtt{modmod}$ outperforms all other methods in terms of learning speed while sharing data reaches the highest final accuracy in less difficult datasets. Federated methods are less effective in modular models; one exception is in the more difficult CIFAR-100 dataset, where it is better to share parameters than data.} 
  \label{fig:modular_lc}
\end{figure}

Fig.~\ref{fig:modular_lc} shows that in all datasets, partial model sharing via $\mathtt{modmod}$ consistently outperforms other methods in the initial stage of learning and provides a massive boost in learning speed. This performance is only eventually topped by $\mathtt{data}$ sharing, once enough data has been shared to characterize the tasks. This validates the claim that optimized partial parameters in the form of modules provide a significant acceleration in learning,  while additional data offer strictly more information not available to the agent otherwise, leading to a higher performance eventually with sufficient training. One notable exception is the more challenging CIFAR-100 dataset where we observe that it is better to share modules than data both in terms of learning speed and final accuracy. Also, we observe that federated methods are less effective in modular networks especially in easier tasks, reaching just roughly the same performance as the isolated baseline in four datasets.

\subsection{How do communication constraints affect the efficacy of sharing modes?} \label{sub:comm_constraints}

\begin{figure}[h!]
\begin{minipage}[b]{.55\textwidth}
  \centering
  \includegraphics[width=\textwidth]{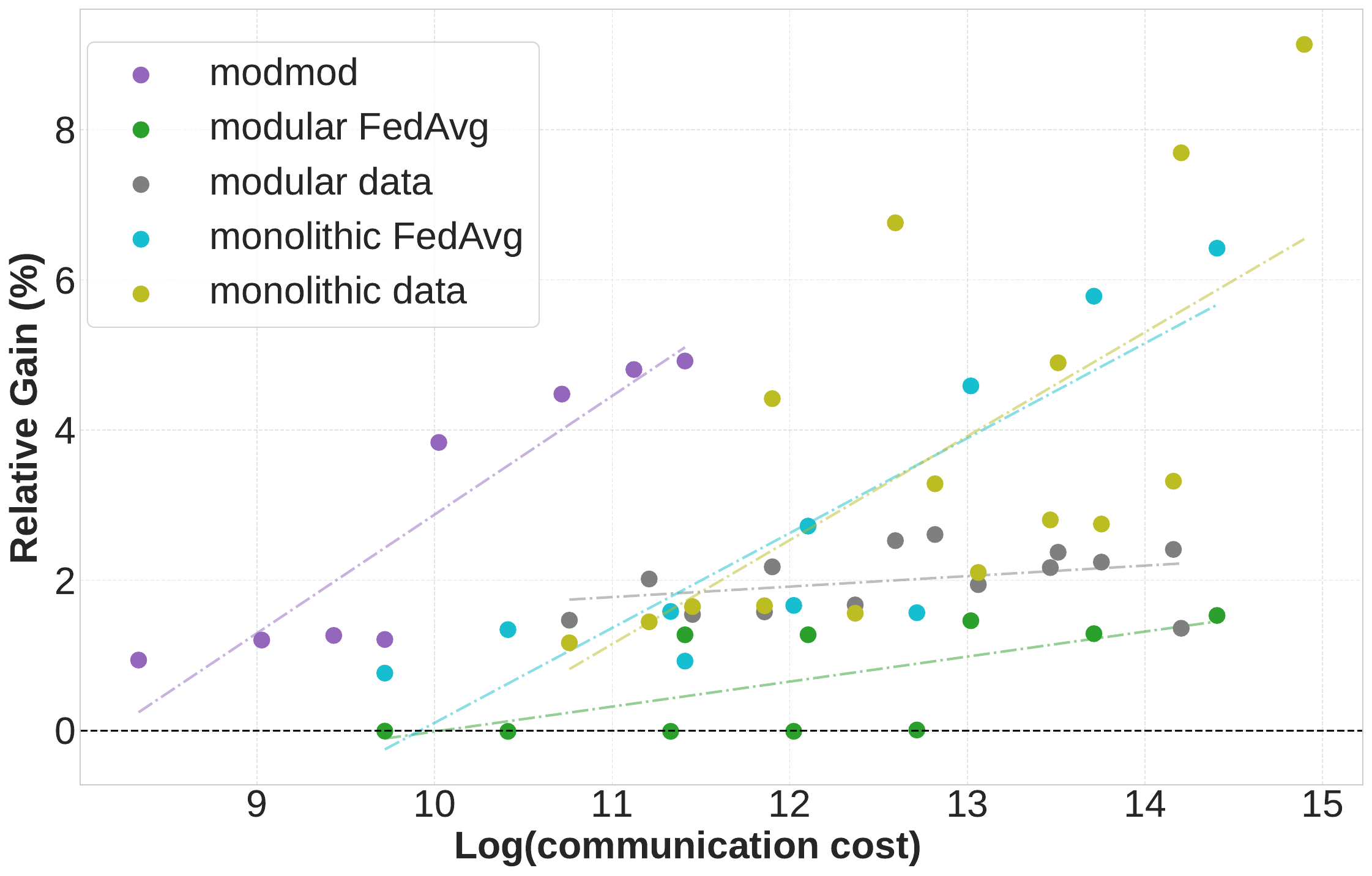}
    \caption{Relative gain in final accuracy versus the log of communication cost,  $\log(\mathtt{B})$.  Lines linearly fit through each sharing mode show the general trend. The results are averaged across agents, random seeds, and datasets. Qualitatively, $\mathtt{modmod}$ improves the final accuracy over isolated learning while requiring substantially less communication than other sharing modes, as further corroborated by Table \ref{tab:marginal_gains_budget}.}
  \label{fig:budget}
\end{minipage}\hfill
\begin{minipage}[b]{.43\textwidth}
\centering
\resizebox{\textwidth}{!}{
\setlength{\tabcolsep}{1pt}
\begin{tabular}{lcc}
\toprule
\textbf{} & \textbf{Marginal} & \textbf{Value of} \\
\textbf{Algorithm} & \textbf{Gain (\%)} & \textbf{Budget} \\
\midrule
ModMod & \textbf{1.579} & $\mathbf{1.17 \!\times\! 10^{-4}}$\\
Modular Federated & 0.332 & $2.56 \!\times\! 10^{-6}$ \\
Modular Data & 0.139 & $9.93 \!\times\! 10^{-6}$ \\
Monolithic Federated & 1.263 & $1.65 \!\times\! 10^{-5}$ \\
Monolithic Data & 1.384 & $1.13 \!\times\! 10^{-5}$ \\
\bottomrule
&\\[2em]
\end{tabular}
}
\captionof{table}{Comparison of marginal gains (higher is better) and value of budget (the ratio of relative gain and communication cost; higher is better) for different sharing modes. $\mathtt{modmod}$ experiences the most favorable scaling with budget (marginal gain) and is an order of magnitude more efficient than other sharing modes (value of budget).}
\label{tab:marginal_gains_budget}
\end{minipage}
\end{figure}

We examine the impact of the communication constraints on the effectiveness of DCL sharing modes.  We vary the number of exchanged data instances, modules, and communication frequency, and record the performance of each sharing mode. Figure \ref{fig:budget} plots the relative gain in the final accuracy of monolithic and modular models over their respective single-agent  baselines against the log of the communication expense $\log(\mathtt{B})$. The communication cost is computed as the total number of floating point numbers exchanged between a pair of agents, $\mathtt{B}=\mathtt{f} \mathtt{b}$. The frequency $\mathtt{f}$ is the number of training epochs before a communication round is initiated, so higher $\mathtt{f}$ implies less frequent communication. $\mathtt{b}$ is the number of floats exchanged per communication. See the Appendix for more details. Table \ref{tab:marginal_gains_budget} computes the marginal gain, which is the slope of the fitted lines in Fig.~\ref{fig:budget}, and the value of the budget, which is the ratio of improvement over the communication cost $\mathtt{B}$. The result shows that partial model sharing via $\mathtt{modmod}$ is an order of magnitude more efficient than other modes.

\subsection{How does network topology affect the efficacy of sharing modes?}

\begin{figure}[h!]
        \centering
        \includegraphics[width=\textwidth]{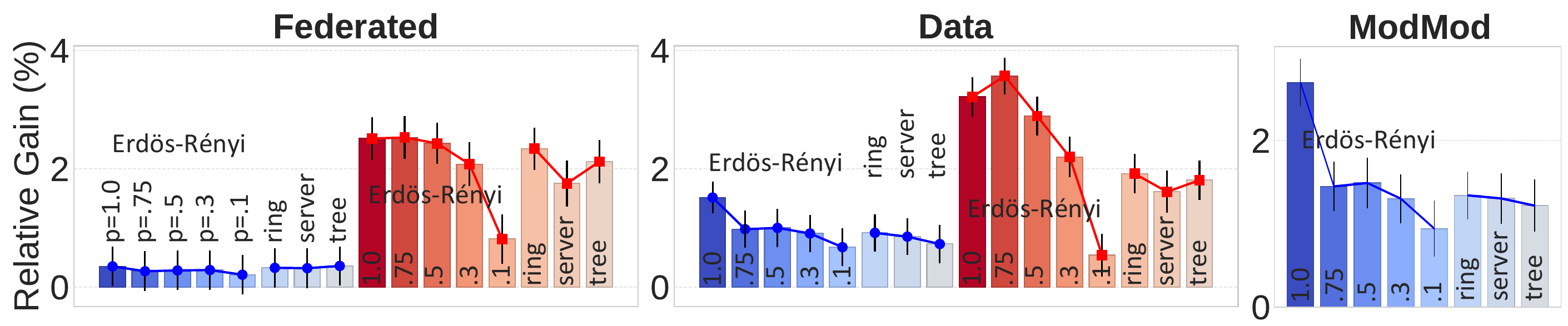}
        \caption{Relative gain in final average accuracy versus topology with \blue{modular} and \red{monolithic} models. We compare four common topologies: Erd\"{o}s-R\'{e}nyi graphs with the probability of each edge $p \in \{1, 0.75, 0.5, 0.3, 0.1\}$, ring, server, and tree. Results are averaged across agents, random seeds, and datasets. On the vertical axis, relative gain = 0\% corresponds to isolated single-agent learning with a disconnected topology ($p=0$). The results show that performance degrades as the graph becomes sparser, i.e. for low values of $p$. Ring, server, and tree have similar performance, falling between a fully connected topology ($p=1$) and the no-sharing baseline. In monolithic neural networks, the server is the worst performing among the three.}
                \label{fig:topology}

\end{figure}

So far, we have only considered a fully-connected communications topology. Now we will examine the effect of network topology on the strategies' performance. We generate Erd\"{o}s–R\'{e}nyi random graphs \cite{Erdos:1959:pmd} with varying degrees of sparsity, and with the ring, server, and tree topologies examined in prior work \cite{rostami2018multiagent}. As shown in Fig.~\ref{fig:topology}, performance generally degrades across all sharing modes as the random graph becomes sparser. Connected topologies such as ring, server, and tree exhibit similar performance, which falls between the fully connected and the no-sharing baseline.

\subsection{What happens when we combine all sharing modes?} \label{sub:combine_modes}

The previous experiments showed that partial model ($\mathtt{modmod}$) and $\mathtt{data}$ sharing offer unique benefits, accelerating learning and improving final accuracy respectively. Here, we examine whether we can have the best of both worlds by combining all sharing modes in one hybrid approach. We compare head-to-head different modes across monolithic and modular networks both in terms of average task area under the curve (AUC) of the learning curve, and final accuracy in Table \ref{tab:all}. The modular hybrid mode strategy consistently outperforms others in both metrics for the more challenging dataset. In easier datasets like MNIST and FashionMNIST, although the performance is more similar across strategies, those involving partial model or data sharing tend to perform best.

\begin{table}[ht!]
\centering
\caption{Performance metrics with std.~error across datasets and algorithms, including a hybrid of all sharing modes. The first number is the final average accuracy (\%) and the second is average AUC.}
\label{tab:all}
\setlength{\tabcolsep}{5pt}
\resizebox{\textwidth}{!}{
\begin{tabular}{llccccc}
\toprule
\textbf{$f$} & \textbf{Method} & \textbf{Cifar100} & \textbf{Combined} & \textbf{Fashionmnist} & \textbf{Kmnist} & \textbf{Mnist} \\
\midrule
\multirow{8}{*}{\rotatebox[origin=c]{90}{modular}}
 & vanilla & \meanstd{70.04}{0.14}\resultssep\meanstd{71.83}{0.13} & \meanstd{88.58}{0.36}\resultssep\meanstd{88.05}{0.28} & \meanstd{92.80}{0.31}\resultssep\meanstd{92.03}{0.32} & \meanstd{80.63}{0.31}\resultssep\meanstd{80.62}{0.31} & \meanstd{93.47}{0.20}\resultssep\meanstd{92.68}{0.22} \\
 & data & \meanstd{71.86}{0.16}\resultssep\meanstd{73.42}{0.14} & \meanstd{89.03}{0.33}\resultssep\meanstd{89.48}{0.25} & \meanstd{93.56}{0.28}\resultssep\meanstd{92.97}{0.30} & \meanstd{82.59}{0.26}\resultssep\meanstd{82.04}{0.29} & \textbf{\meanstd{94.60}{0.12}}\resultssep\meanstd{93.89}{0.14} \\
 & fedavg & \meanstd{70.99}{0.14}\resultssep\meanstd{73.96}{0.14} & \meanstd{88.45}{0.37}\resultssep\meanstd{87.99}{0.31} & \meanstd{92.46}{0.45}\resultssep\meanstd{91.96}{0.45} & \meanstd{80.04}{0.35}\resultssep\meanstd{80.10}{0.34} & \meanstd{93.57}{0.18}\resultssep\meanstd{92.82}{0.23} \\
 & fedfish & \meanstd{71.10}{0.13}\resultssep\meanstd{73.97}{0.15} & \meanstd{88.46}{0.37}\resultssep\meanstd{87.99}{0.29} & \meanstd{93.08}{0.39}\resultssep\meanstd{92.33}{0.44} & \meanstd{80.32}{0.36}\resultssep\meanstd{80.13}{0.38} & \meanstd{93.77}{0.19}\resultssep\meanstd{92.76}{0.26} \\
 & fedcurv & \meanstd{71.11}{0.13}\resultssep\meanstd{73.94}{0.14} & \meanstd{88.45}{0.37}\resultssep\meanstd{87.99}{0.29} & \meanstd{93.06}{0.40}\resultssep\meanstd{92.32}{0.44} & \meanstd{80.34}{0.36}\resultssep\meanstd{80.14}{0.38} & \meanstd{93.78}{0.19}\resultssep\meanstd{92.76}{0.26} \\
 & fedprox & \meanstd{71.82}{0.14}\resultssep\meanstd{73.85}{0.13} & \meanstd{88.47}{0.37}\resultssep\meanstd{88.02}{0.29} & \meanstd{92.93}{0.41}\resultssep\meanstd{92.07}{0.46} & \meanstd{80.48}{0.35}\resultssep\meanstd{80.20}{0.35} & \meanstd{93.93}{0.17}\resultssep\meanstd{93.05}{0.21} \\
 & modmod & \meanstd{76.77}{0.08}\resultssep\meanstd{76.80}{0.13} & \meanstd{89.65}{0.30}\resultssep\meanstd{89.62}{0.26} & \meanstd{93.28}{0.40}\resultssep\meanstd{93.26}{0.41} & \meanstd{81.83}{0.27}\resultssep\meanstd{81.78}{0.29} & \meanstd{94.08}{0.18}\resultssep\meanstd{93.94}{0.19} \\
 & hybrid & \textbf{\meanstd{77.05}{0.10}}\resultssep\textbf{\meanstd{77.07}{0.14}} & \textbf{\meanstd{90.31}{0.28}}\resultssep\textbf{\meanstd{90.41}{0.24}} & {\meanstd{94.10}{0.31}}\resultssep\meanstd{94.43}{0.35} & \textbf{\meanstd{82.93}{0.23}}\resultssep\textbf{\meanstd{82.90}{0.25}} & \meanstd{94.57}{0.12}\resultssep{\meanstd{94.75}{0.14}} \\\hline
\multirow{6}{*}{\rotatebox[origin=c]{90}{monolithic}} & vanilla & \meanstd{63.11}{0.32}\resultssep\meanstd{65.60}{0.32} & \meanstd{86.59}{0.37}\resultssep\meanstd{87.69}{0.28} & \meanstd{92.47}{0.33}\resultssep\meanstd{93.27}{0.36} & \meanstd{78.23}{0.31}\resultssep\meanstd{79.42}{0.31} & \meanstd{91.96}{0.27}\resultssep\meanstd{93.10}{0.22} \\
 & data & \meanstd{65.18}{0.17}\resultssep\meanstd{67.23}{0.21} & \meanstd{87.82}{0.36}\resultssep\meanstd{88.59}{0.28} & \meanstd{94.02}{0.35}\resultssep{\meanstd{94.72}{0.38}} & \meanstd{80.70}{0.32}\resultssep\meanstd{81.01}{0.29} & \meanstd{94.20}{0.17}\resultssep\meanstd{94.54}{0.21} \\
 & fedavg & \meanstd{67.05}{0.16}\resultssep\meanstd{69.11}{0.18} & \meanstd{86.57}{0.40}\resultssep\meanstd{86.76}{0.31} & \meanstd{93.68}{0.39}\resultssep\meanstd{93.60}{0.39} & \meanstd{80.55}{0.33}\resultssep\meanstd{80.49}{0.35} & \meanstd{93.86}{0.20}\resultssep\meanstd{93.76}{0.26} \\
 & fedfish & \meanstd{66.85}{0.17}\resultssep\meanstd{68.91}{0.17} & \meanstd{86.55}{0.41}\resultssep\meanstd{86.82}{0.31} & \meanstd{93.64}{0.38}\resultssep\meanstd{93.84}{0.40} & \meanstd{80.39}{0.34}\resultssep\meanstd{80.38}{0.36} & \meanstd{93.65}{0.24}\resultssep\meanstd{93.37}{0.31} \\
 & fedcurv & \meanstd{66.79}{0.15}\resultssep\meanstd{68.93}{0.17} & \meanstd{86.62}{0.41}\resultssep\meanstd{86.87}{0.31} & \meanstd{93.74}{0.37}\resultssep\meanstd{93.85}{0.40} & \meanstd{80.39}{0.34}\resultssep\meanstd{80.40}{0.35} & \meanstd{93.66}{0.24}\resultssep\meanstd{93.37}{0.31} \\
 & fedprox & \meanstd{67.18}{0.17}\resultssep\meanstd{69.09}{0.16} & \meanstd{86.86}{0.40}\resultssep\meanstd{87.16}{0.31} & \meanstd{93.90}{0.35}\resultssep\meanstd{93.99}{0.39} & \meanstd{80.26}{0.33}\resultssep\meanstd{80.47}{0.35} & \meanstd{93.74}{0.23}\resultssep\meanstd{93.65}{0.30} \\
  & hybrid & \meanstd{70.05}{0.17}\resultssep\meanstd{69.15}{0.16} & \meanstd{88.28}{0.36}\resultssep\meanstd{88.76}{0.29} & \textbf{\meanstd{94.66}{0.32}}\resultssep\textbf{\meanstd{94.97}{0.35}} & \meanstd{81.06}{0.31}\resultssep\meanstd{81.01}{0.30} & \meanstd{94.55}{0.20}\resultssep\textbf{\meanstd{94.90}{0.21}} \\

\bottomrule
\end{tabular}
}
\end{table}

\section{Conclusion}
We have developed a rigorous and general definition for the DCL problem and investigated a key under-explored challenge: determining at which level to share information between agents, from the data level over partial models (modules) to full models. Our results show that partial model sharing in the form of reusable modules achieves high performance (due to reusability and flexibility) with low communications cost. When combined with data sharing in a hybrid approach, the combined modalities achieve even higher performance. By providing more competitive sharing baselines and drawing attention to the hidden cost of communication as well as assumptions on agent commonality, we address the shortcomings of earlier works and enable more robust evaluation. While our work establishes a solid bases for future DCL research, one limitation is its current focus on supervised learning. In the future, we aim to extend our work to other fields such as reinforcement learning.

\section*{Acknowledgement}
This research was partially supported by the DARPA SHELL program under contract HR0011-21-9-0133, the Army Research Office MURI W911NF20-1-0080, and the DARPA Triage Challenge under award HR001123S0011.

\bibliography{references}
\bibliographystyle{unsrt}  

\newpage
\appendix

\begin{center}
    {\Large Technical Appendices for ``Distributed Continual Learning''}
\end{center}

\section{Additional Experimentation Details}

\paragraph{Datasets} We use four common datasets MNIST, KMNIST, FashionMNIST, and CIFAR-100), following the same task sampling procedure from prior work \cite{mendez2021lifelong} to create multiple tasks. The MNIST variants consist of 10 tasks while CIFAR has 20 tasks, where each task is a multiclass classification problem. Each agent receives different tasks drawn from the same dataset. $\mathtt{combined}$ setting is created by including tasks from MNIST, KMNIST, and FashionMNIST. In this setting, for a few (four) initial tasks, all agents receive a mixture of tasks from the three source datasets; then, they segregate into three roughly equal-sized groups, each receiving tasks drawing from exactly one source dataset. The $\mathtt{combined}$ network has 20 agents while other datasets have eight agents.

\paragraph{Network architectures} The agent's models are multilayer perceptrons (MLPs) for MNIST variants and convolutional neural networks (CNNs) for CIFAR. Following prior work \cite{mendez2021lifelong}, both monolithic and modular networks start out with four modules but modular networks are allowed to dynamically add more modules. 

\paragraph{Experimental Protocols} All experiments are repeated with eight random seeds.

\paragraph{Communication Budget} In the basic experiments (Sec. \ref{sub:monolithic}-\ref{sub:modularExps}), each agent in data sharing is allowed $20$ queries with $5$ query neighbors every $16$ epochs while federated methods communicate every $5$ epochs, and $\mathtt{modmod}$ communicates once every task. In an extensive experiment (Sec. \ref{sub:comm_constraints}), we investigate the effect of varying the frequency of communication, $\mathtt{f}$, and the budget per communication, $\mathtt{b}$. For federated methods, we set $\mathtt{f}$ = 5, 10, 20, 50, 100 epochs. For $\mathtt{modmod}$, we exchange modules once per task and vary the number of sent modules (effectively $\mathtt{b}$) in 1, 2, 3, 4. For data sharing, we vary both the number of queries in $\mathtt{q}$ = 10, 20, 30 and the communication frequency in $\mathtt{f}$ = 9, 16, 50.

\paragraph{Sharing Algorithms} For data sharing, we use $\mathtt{Recv}$ for MNIST variants and $\mathtt{combined}$, and use $\mathtt{Simp}$ for CIFAR-100. This is because $\mathtt{Recv}$ is most effective when the sender can accurately retrieve relevant instances to a query; this is hard for CIFAR-100, so to ablate away the deficiency of the learners in $\mathtt{Recv}$ and to give data sharing a fair chance against other modes, we instead use the non-learning $\mathtt{Simp}$. For full model parameter sharing, we implement $\mathtt{FedAvg}$, $\mathtt{FedProx}$, $\mathtt{FedCurv}$, and $\mathtt{FedFish}$. As described in Sec.~\ref{sub:modular}, modular neural networks support the dynamic addition of modules, thus agents might have different number of modules at any given time. So for federated methods, we only aggregate the basis modules---modules shared by all agents that are created during initialization \cite{mendez2021lifelong}. The number of basis modules is set to four following the previous work. For $\mathtt{modmod}$, we use use $\mathtt{LEEP}$ metric for $\mathtt{combined}$ dataset and $\mathtt{IoU}$ for the others. $\mathtt{TrustSim}$ module selection is used in the basic experiment (Sec. \ref{sub:monolithic}-\ref{sub:modularExps}) but as we get more candidate modules in the increasing budget (Sec. \ref{sub:comm_constraints}) we use $\mathtt{TryOut}$ instead. 

\paragraph{Topology}
Erd\"{o}s–R\'{e}nyi random graphs \cite{Erdos:1959:pmd} with varying degrees of sparsity, and with the ring, server, and tree topologies examined in prior work \cite{rostami2018multiagent} are used.

\paragraph{Compute Resources} We use Ray distributed library to parallelize agent training in a communication network. One single NVIDIA GeForce RTX 3080 takes about 10 minutes to train MNIST and $\mathtt{combined}$ datasets, and 1 hour for CIFAR-100.

\paragraph{Hyper-parameters} FedProx and FedCurv algorithms are run with the loss balancing factor $\mu \in \{0.001, 0.01, 0.1, 1.0\}$, following \cite{ li2020federated,shoham2019overcoming}, and the best performing hyper-parameter for each dataset in terms of average AUC is picked for the paper.

\section{Budget Computation}
In Sec.~\ref{sub:comm_constraints}, we compare different sharing modes in terms of their use of $\mathtt{b}$, the number of floats exchanged per communication. We now provide more details on how $\mathtt{b}$ is computed.

For data sharing, each instance is an image, represented in memory as a tensor of size $H \cdot W \cdot C$ for height $H$, width $W$, and number of channels $C$. Then, $\mathtt{b} = H \cdot W \cdot C \cdot N$ where $N$ is the number of instances sent per communication.

For full model parameters sharing, a monolithic neural network consists of some feed-forward or convolution layers followed by a last linear layer mapping to the semantic classes. The number of parameters of a linear layer is $|W| + |b|$ for some weight matrix $W$ and bias vector $b$. The number of parameters of a convolution layer is $(c_{\textrm{in}}c_{\textrm{out}}) k^2 + c_{\textrm{out}}$ where $c_{\textrm{in}},c_{\textrm{out}}$ are in and out channels and $k$ is the kernel size. The communication cost $\mathtt{b}$ is then simply the size of the model. Although not investigated in this work, a future direction is to use model compression such as low-rank factorization \cite{vogels2020powergossip} to have a finer-grain control over the cost of sending models in federated methods. 

For modular parameter sharing, $\mathtt{b} = k \cdot M$ where $k$ is the number of shared modules, and $M$ is the module size in terms of parameters computed analogously as explained above.

\section{Additional Results on Hybrid Sharing}

\begin{figure}[h!]
    \centering
    \begin{subfigure}{\textwidth}
        \centering
        \includegraphics[width=0.9\textwidth]{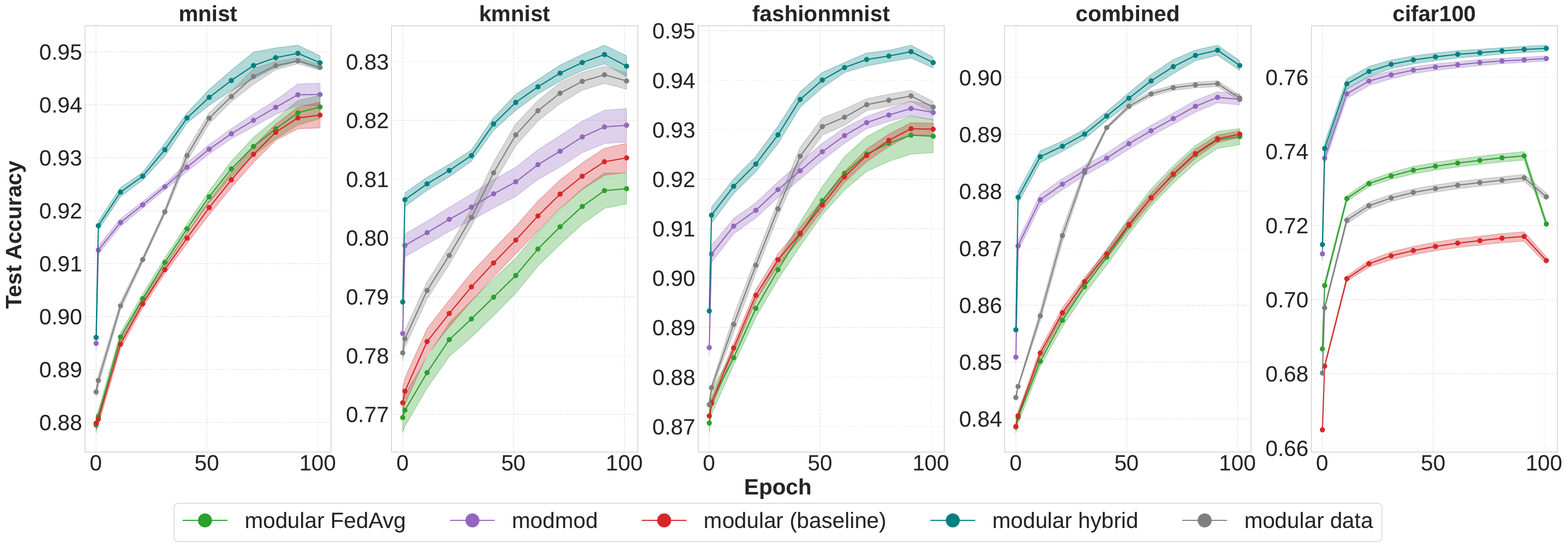}
        \label{fig:combine_modes_modular_lc}
    \end{subfigure}%
    
    \begin{subfigure}{\textwidth}
        \centering
        \includegraphics[width=0.9\textwidth]{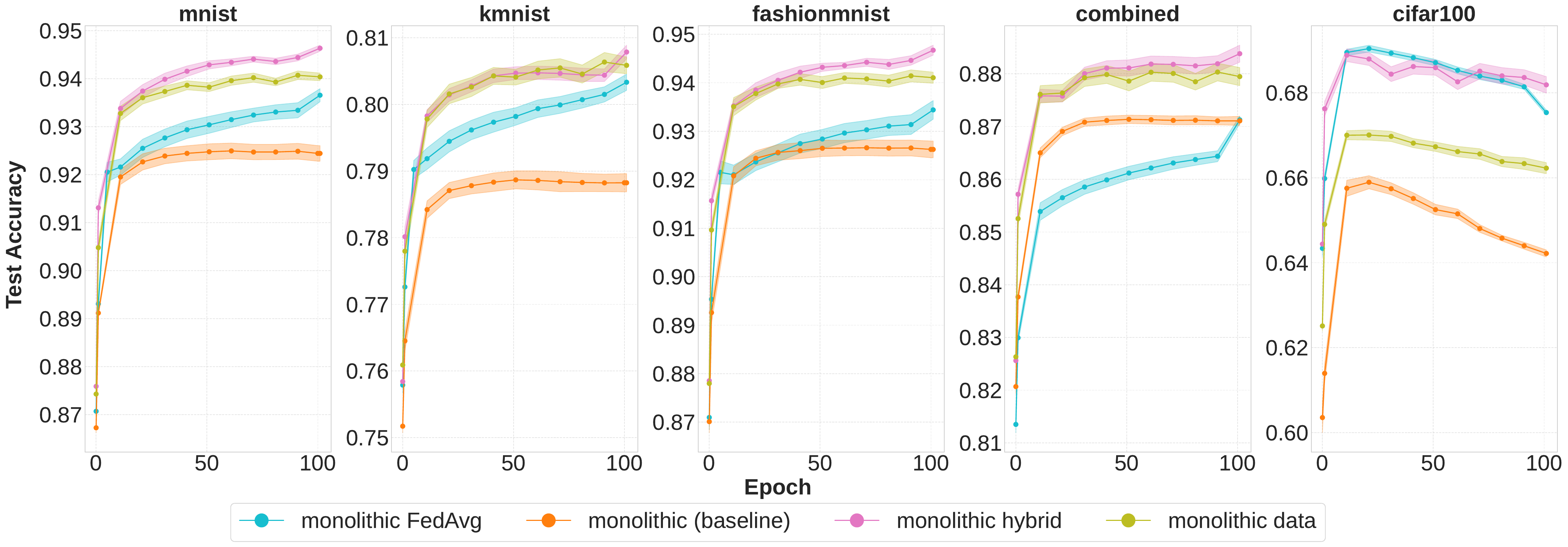}
        \label{fig:combine_modes_monolithic_lc}
    \end{subfigure}
    \caption{Mean test accuracy and standard error of a hybrid combination of all sharing modes compared against individual modes in modular (top) and monolithic (bottom) networks. In modular networks, hybrid mode achieves the best performance both in learning speed and final accuracy. The superiority of hybrid mode is less observed in monolithic networks.}
    \label{fig:combine_modes}
\end{figure}

The results in this section complement those in Section~\ref{sub:combine_modes}. Fig.~\ref{fig:combine_modes} shows the learning curves for the hybrid sharing, combining all sharing modes, against the other modes. It supports the results in Table~\ref{tab:all} and shows that by combining all sharing modes in modular networks, we achieve both good initial learning and final accuracy. In monolithic networks, the hybrid mode generally outperforms others although the gain is less pronounced.

The modular hybrid mode strategy consistently outperforms others in both metrics for the more challenging dataset. In easier datasets like MNIST and FashionMNIST, although the performance is more similar across strategies, those involving $\mathtt{modmod}$ or data tend to perform best. 

\begin{table}[H]
\centering
\caption{Relative Performance Gaps for Final and AUC Scores across Datasets. These performance gaps are reasonable proxies for task difficulties. We observe substantial improvement of DCL over single-agent learning in difficult tasks like cifar100 while the gap in performance between various strategies in easier tasks like fashionmnist is smaller.}
\label{tab:relative_gaps_percentage}
\begin{tabular}{lcc}
\toprule
\textbf{Dataset} & \textbf{Final Gap (\%) } & \textbf{AUC Gap (\%) } \\
\midrule
cifar100 & 22.09\% & 17.50\% \\
combined & 4.33\% & 4.20\% \\
kmnist & 6.02\% & 4.38\% \\
fashionmnist & 1.78\% & 3.00\% \\
mnist & 2.87\% & 2.23\% \\
\bottomrule
\end{tabular}
\label{tab:gap}
\end{table}

\end{document}